\documentclass{article} 
\usepackage[final]{colm2025_conference}

\usepackage{microtype}
\usepackage{hyperref}
\usepackage{url}

\usepackage[TABBOTCAP]{subfigure}
\usepackage[shortlabels]{enumitem}
\usepackage{placeins}
\usepackage{bbold}
\usepackage{tikz-dependency}
\usepackage{algorithm}
\usepackage{algpseudocode}
\usepackage{multirow}
\usepackage{booktabs}
\usepackage{soul}
\usepackage{color}
\usepackage{colortbl}
\usepackage[many]{tcolorbox}
\usepackage{lineno}

\definecolor{darkblue}{rgb}{0, 0, 0.5}
\hypersetup{colorlinks=true, citecolor=darkblue, linkcolor=darkblue, urlcolor=darkblue}

\usepackage{xcolor}
\usepackage{color-edits}
\addauthor{hm}{blue}
\addauthor{sw}{orange}
\addauthor{xr}{cyan}
\addauthor{tc}{magenta}
\addauthor{sc}{purple}

\title{The Ram\'{o}n Llull's Thinking Machine for \\Automated Ideation}

\author{Xinran Zhao$^1$\quad
 \textbf{Boyuan Zheng$^{\dagger2}$}\quad
 \textbf{Chenglei Si$^{\dagger3}$}\quad
 \textbf{Haofei Yu$^{\dagger4}$}\quad
  \textbf{Ken Ziyu Liu$^{\dagger3}$}\quad\\
 \textbf{Runlong Zhou$^{\dagger6}$}\quad
 \textbf{Ruochen Li$^{\dagger5}$}\quad
 \textbf{Tong Chen$^{\dagger6}$}\quad
 \textbf{Xiang Li$^{\dagger4}$}\quad
 \textbf{Yiming Zhang$^{\dagger1}$}\quad\\
  \textbf{Tongshuang Wu$^1$}~\thanks{\quad $^{\dagger}$ denotes equal contribution in alphabetic order. Corresponding contact email addresses: \{xinranz3,sherryw\}@andrew.cmu.edu.} \\
 $^1$CMU, $^2$OSU, $^3$Stanford, $^4$UIUC, $^5$UT Dallas $^6$UW
}

\begin{document}

\ifcolmsubmission
\linenumbers
\fi

\maketitle

\begin{abstract}

This paper revisits Ram\'{o}n Llull’s \textit{Ars combinatoria}---a medieval framework for generating knowledge through symbolic recombination---as a conceptual foundation for building a modern Llull's ``\textit{thinking machine}'' for research ideation.
Our approach defines three compositional axes: Theme (\textit{e.g.}, efficiency, adaptivity), Domain (\textit{e.g.}, question answering, machine translation), and Method (\textit{e.g.}, adversarial training, linear attention). These elements represent high-level abstractions common in scientific work---motivations, problem settings, and technical approaches---and serve as building blocks for LLM-driven exploration. We mine elements from human experts or conference papers and show that prompting LLMs with curated combinations produces research ideas that are diverse, relevant, and grounded in current literature. This modern thinking machine offers a lightweight, interpretable tool for augmenting scientific creativity and suggests a path toward collaborative ideation between humans and AI.

\end{abstract}

\section{Introduction}
\label{sec:introduction}

There is a growing interest in the machine learning community in leveraging large language models (LLMs) to accelerate scientific discovery~\citep{si2024llmideas, AI4Science2023TheIO, collins2024evaluating, singh2025ai2scholarqaorganized, jansen2025codescientistendtoendsemiautomatedscientific,si2025ideationexecutiongapexecutionoutcomes}. Among these prominent directions, one challenging topic is to use LLMs to conduct or assist the \textit{ideation} process. 
Despite recent success in ideation with state-of-the-art language models~\citep{si2024llmideas}, community simulation~\citep{yu2024researchtown}, and reinforcement learning~\citep{li2024learning}, model-generated ideas can lack diversity~\citep{si2024llmideas}. In response, in this paper, we ask: 
\textit{Does conditioning on explicit concept combinations help build a minimalist pipeline for diverse and grounded research ideas?}

An ideal pipeline shall be simple, scalable, and it can generate a diverse set of ideas. In this work, we propose to create such a pipeline through revisiting one of the first human explorations of artificial intelligence invented at the end of the thirteenth century, which aims at creating new knowledge from logical combinations of concepts~\citep{borges1937}. Llull's machine includes multiple rotary disks of concepts, \emph{e.g.}, goodness, power, glory, etc, where Llull believed studying all combinations of the elementary concepts would help understand a field of knowledge that can be covered by them. In light of the thinking, we revisit the idea of element combination to create a modern version for LLM ideation. 

Specifically, we design three \textit{disks} of elements \textit{theme}, \textit{domain}, and \textit{method}. Corresponding research ideas are then synthesized with a finite set of rules combining all elements\footnote{Such a categorization is not exhaustive. We discuss this in the limitations section in the appendix.}. For example, with \textit{less is more} as a \textit{theme}, \textit{confidence calibration} as a \textit{domain},  \textit{Mamba}~\citep{gu2024mamba} as a \textit{method}, and a simplest \textit{A+B+C} template, after rewriting the raw idea with Claude 3.7~\citep{anthropic2024claude}, a candidate idea can be: \textit{Less Parameters, Better Calibration: Confidence-Aware Training for Mamba Architectures}. We conduct a pilot study validating the pipeline with human-written elements and then scale the ideation with elements mined automatically from top-tier conferences, \emph{e.g.}, ICLR, ACL, etc. 

To study the characteristics of the ideas, we first compare the statistics and elements (themes, domains, and methods) extracted from different conferences across years, which sheds light on the taste and preferences of different machine learning communities, \emph{e.g.}, from the same number of papers, our pipeline extracts similar numbers of domain elements from ACL and more method elements from ICLR. Next, with the raw ideas combined through automatically extracted templates in the same pipeline, we further use LLMs to rewrite them into research ideas. We compare these output ideas from Ram\'{o}n Llull's Thinking Machine with idea titles from previous work~\citep{si2024llmideas, yu2024researchtown}, which suggests good diversity and coverage of the ideas generated from our minimalist method\footnote{The authors note that the diversity and coverage do not necessarily suggest the novelty and utility of the ideas, which require extensive human experimentation and evaluation to validate.}.

In this paper, we explore the potential of LLM ideation through reconstructing the thirteenth-century Ram\'{o}n Llull's thinking machine with modern data mining and automatic evaluation techniques. We anticipate the proposed pipeline and resources to serve as (1) a simple but strong baseline for LLM ideation; (2) an interesting view that motivates human researchers to find or review their ideas. The authors acknowledge the core contribution of our work as an investigation into quantifying how much research ideation can be mechanically automated.
We will open-source our code, data, and generated research ideas at \url{https://github.com/colinzhaoust/ramon_llull_public}.

\section{Related Work}
\label{sec:related_work}

\paragraph{Symbolic Reasoning} Ram\'{o}n Llull's thinking machine~\citep{borges1937} is one of the earliest attempts at formalizing reasoning, laying the foundation for symbolic AI. It motivates later developments such as mathematized logic~\citep{uckelman2010computing}, the universal Turing machine~\citep{turing1universal}, ontologies~\citep{goerss2024mirror} and knowledge graphs~\citep{ji2021survey}. While it shares with knowledge graphs the goal of representing structured information, the key difference lies in their operational principles. Knowledge graphs capture large-scale relational structures among extracted entities. 

In contrast, Llull's thinking machine starts with a small and fixed set of core concepts and systematically explores their combinatorial possibilities using a rotating mechanism. This generative, combinatorial focus distinguishes it from the more static and structural nature of knowledge graphs.

\paragraph{Automatic Ideation}
Recent advancements have explored the use of LLMs to automate and enhance scientific and creative ideation. The most direct approach involves prompting LLMs to generate ideas in a single pass~\citep{si2024llmideas}. Building on this, other works incorporate more structured techniques such as iterative boosting~\citep{Wang2023SciMONSI},  knowledge augmentation ~\citep{baek2024researchagent}, multi-agent collaboration~\citep{yu2024researchtown}, reinforcement learning~\citep{li2024learning}, to refine ideation quality. A further step involves analogical reasoning~\citep{hope2017accelerating}, which mines high-quality ideas from structured knowledge by drawing connections between similar concepts.

Our approach moves one step beyond analogy: we identify high-quality core concepts and systematically explore their combinatorial space—inspired by Llull’s thinking machine—to generate novel and diverse ideas grounded in specific research communities.

Concurrently, Scideator~\citep{radensky2024scideator} recombines facets extracted from papers on different dimensions for ideation. CHIMERA~\citep{sternlicht2025chimera} provides a knowledge base of linked recombination mined from papers from various domain.

We further discuss related work on data mining from academic papers in Appendix~\ref{appendix:extended_discussion}.

\section{The Ram\'{o}n Llull's Thinking Machine}
\label{sec:methodology}
From the historical context, Ram\'{o}n Llull designed the machine to provide answers to arbitrary questions with a combination of elements selected through spinning three concentric and revolving wood or metal disks\footnote{We present a figurative illustration in Appendix~\ref{fig:original_machine}}. Through patient manipulation of the multiplication and elimination, the machine will eventually produce a seemingly good answer.

We consider this process a simulation of one kind of human ideation process: a researcher may see a good paper and decide to apply it to their own domains, \emph{e.g.}, the recent success in introducing teacher forcing to diffusion~\citep{chen2024diffusion}\footnote{We also note other processes, \emph{e.g.}, see an abnormal phenomenon~\citep{goyal2025contextparametric}, answer a question of the community~\citep{wu2024replymakelovetal}, and push to the extreme condition~\citep{shao2024scaling}.}. There is no guarantee on if this ideation process is the best in terms of novelty, but it shall be considered as a common practice in various communities.

\begin{figure}
    \centering
     \includegraphics[clip,trim={0cm 0cm 0cm 0cm},width=\linewidth]{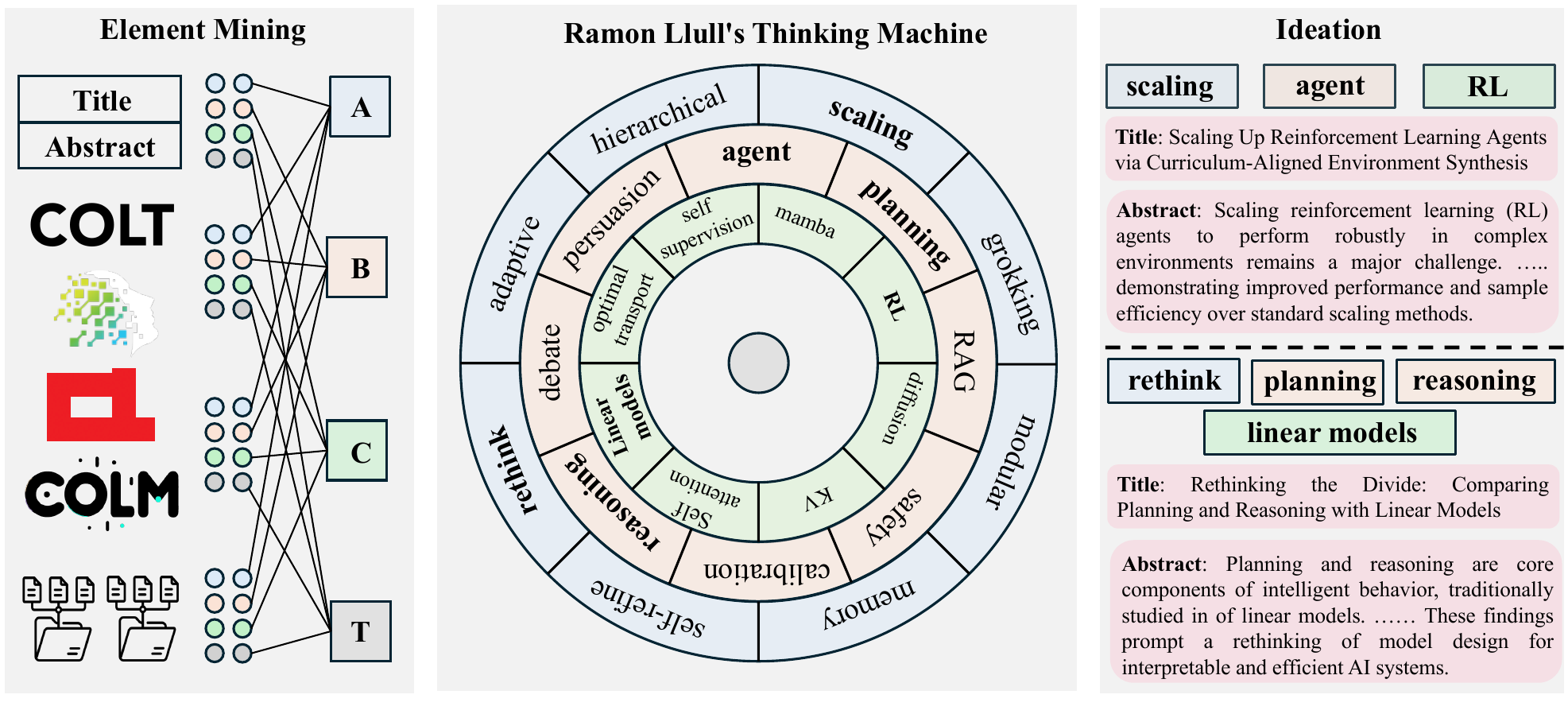}
    \caption{The overall pipeline of using the concept of Ram\'{o}n Llull's thinking machine for research ideation. It includes three main steps (1) \textit{element mining}: mining and merge elements (like keywords representing themes, domains, and methods) from papers in top conferences; (2) \textit{combinational thinking}: combining extracted elements through symbolic recombination similar with Ram\'{o}n Llull's thinking machine; (3) \textit{idea generation}: generating abstract-style research ideas based on templates and elements.}
    \label{fig:pipeline}
\end{figure}

In a formal way, given three disks of elements $A=\{a_1, a_2,...\}, B=\{b_1, b_2,...\}, C=\{c_1, c_2,...\}$ and a template $T$~\footnote{For example, $a+b+c$ or compare $c_1$ and $c_2$ in $b_1$ under $a_1$}, Ram\'{o}n Llull's Thinking Machine $\phi$ outputs the raw idea $x = \phi(A, B, C, T)$, where $T$ can require $\geq1$ elements from each disk. Then, given a large language model $m$, following the setup of~\citep{si2024llmideas}, we denote the ideation as the generation of a title and a corresponding abstract, $(t,{abs}) \sim m(x)$. We show our pipeline of element mining and ideation in Figure~\ref{fig:pipeline}, with details in the following sections. At this stage, we leave sampling execution plans from the raw idea to future work.

\subsection{Building the Idea Generator}

\paragraph{Elements.} Similar to the original thinking machine, we design three disks to capture the minimum description of the ideation context:  

\begin{itemize}[labelwidth=*,leftmargin=1.3em,align=left, topsep=0pt, nosep]
\item Theme (\textit{A}): the theme of the work, which highlights a particular scene or purpose to conduct the study, \emph{e.g.}, \textit{less is more}, \textit{few-shot}, \textit{adaptive}, \textit{aggregation}, \textit{in-the-wild}, \textit{is all you need}, etc. The theme elements might be revisited with different names in the literature, where ``trendy'' themes can also be different across communities.

\item Domain (\textit{B}): the domain of the work, which indicates a potential set of tasks to solve and previous work to follow, \emph{e.g.}, \textit{argument mining}, \textit{question answering}, etc.

\item Method (\textit{C}): the method of the work, which shows how certain problems are addressed with specific adoption or adaptation of a model, data, or training framework, \emph{e.g.}, \textit{transformer}, \textit{state-space models}, \textit{preference optimization}, etc.
\end{itemize}

We select the current disks to form the minimum description of a research idea: we did $a$ in $b$ with $c$, as described in the IMRaD format of academic writing~\citep{imrad}. However, elements in the disks can be non-exclusive, \emph{e.g.}, \textit{retrieval} can be considered as a domain with various tasks, as well as a set of methods for other domains. We discuss other potential axes of the machine in Section~\ref{sec:beyond_abc}. We further discuss the relations of the intra/inter-disk elements in Appendix~\ref{appendix:limitations}.

\paragraph{Ideation.} With the disks in hand, to capture the diverse relations and combinations among elements, we extend the original Ram\'{o}n Llull's Thinking Machine with templates for generation, which is widely used in the knowledge graph construction~\citep{ZhangLPSL20}. A template serves as a way to combine the elements, with potential additional descriptive words on their relations. Besides the aforementioned ``we did $a$ in $b$ with $c$'', other rudimentary templates can be ``compare $c_1$ and $c_2$ in $b_1$ under $a_1$'' or ``$c_1$ is all you need''.


\subsection{Mining the elements and templates}
\label{sec:elements}

\begin{table}[t]
\centering
\small
\begin{tabular}{p{1.8cm}|p{4cm}|p{3cm}|p{3cm}}
\toprule
\textbf{Community} & \textbf{A (Theme)} & \textbf{B (Domain)} & \textbf{C (Method)} \\
\midrule
NLP & adaptive, less is more, hierarchical, in-the-wild, self-refine, hindsight, rethink, grokking, long-tail, compositional, multi-hop  & agent, planning, retrieval, safety, calibration, reasoning, memorization, persuasion, debate & Mamba, RL, Linear Models, KV Cache, Quantization, Diffusion, Self-attention, Self-supervision \\
\midrule
CV & test-time Training, meta learning, active learning, open-set calibration, open-vocab grounding, continual learning & image classification, detection, segmentation, optical flow estimation, action recognition & ViT, NeRF, ConvNext, point-transformer, Perceiver, Instant-NGP, Yolo, UNet, LoRA\\
\midrule
RL Theory & parametric policy optimization, online learning, offline learning, adversarial, corruption, linear policy, general function approximation & multi-armed / contextual bandits, Markov decision processes, Markov games, stochastic shortest path  & $\varepsilon$-greedy, Thompson sampling, upper confidence bound, optimism, pessimism  \\
\bottomrule
\end{tabular}
\caption{\small{Lists of Theme (A), Domain (B), and Method (C) written by researchers from different communities. NLP, CV, and RL Theory denote natural language processing, computer vision, and reinforcement learning theory, respectively. We present the full table in Appendix~\ref{appendix:elements}.}}
\label{tab:human_abc}
\end{table}

\begin{table}[t]
\centering
\small
\begin{tabular}{l|c|c|c|c|c|c|c}
\toprule
\textbf{Stats.} & ICLR 24 & COLM 24 & COLT 24 & ACL 24 & ACL 23 & ACL 22  & All\\
\midrule
\# Papers  & 2000 & 299 & 170 & 1931 & 2052 & 1031 & 7483  \\
\midrule
\# Theme (A) & 391 & 118 & 91 & 307 & 359 & 224 & 682 \\
\midrule
\# Domain (B) & 330 & 87 & 62 & 300 & 272 & 208 & 633 \\
\midrule
\# Method (C) & 392 & 35 & 53 & 54 & 117 & 153 & 866 \\
\midrule
\#Template (T) & 278 & 71 & 75 & 121 & 277 & 165 & 925 \\
\bottomrule
\end{tabular}
\caption{Statistics of the papers processed: themes, domains,  methods, and templates mined from various top-tier conferences. \textit{All} denotes the cumulative elements after merging and filtering. \# X denotes the number of X.}
\label{tab:conference_stats}
\end{table}

\paragraph{Pilot: Human Annotation.}
To validate our design of the disks, we first seek elements of the disks from PhD students from different communities\footnote{Each volunteer has published 5+ papers in the conferences of the corresponding community.}. Table~\ref{tab:human_abc} presents the theme, domain, and method elements written by human experts. With LLM rewriting, these elements can already lead to interesting research ideas, \emph{e.g.}, from \textit{hindsight, debate, RL}, Claude 3.7 can output a title: \textit{Learning to Argue in Hindsight: Multi-Agent Debate with Retrospective Reinforcement Learning} with a reasonable abstract.

Among different communities, there are similarities, \emph{e.g.}, \textit{transformer} and \textit{diffusion}, and differences, \emph{e.g.}, \textit{multi-hop} vs. \textit{inverse rendering}. In recent years, certain ideas from one community have motivated the novel directions in other communities, \emph{e.g.}, transformer for vision~\citep{dosovitskiy2021an} and diffusion for text~\citep{Li-2022-DiffusionLM}, and vice versa, which inspires us to study and auto-extraction of these elements from conferences acknowledged in different communities.

\paragraph{Mining from the Literature.} To automate and scale the element harvest, we propose to automatically mine the elements and templates from different top-tier conferences, which allows for extendability to our pipeline. Specifically, we use Gemini 2.0 Flash~\citep{geminiteam2024gemini15unlockingmultimodal} to process each paper title and abstract into lists of A, B, C, and a template with a carefully designed element extraction prompt. Upon acquiring the elements from different papers, since we observe duplications among the elements, we then leverage Gemini 2.0 Flash again to merge the elements based on the semantic similarities. The detailed prompts are presented in Appendix~\ref{appendix:experimental_details}.

We collected the papers from Paper Copilot~\citep{Yang2025PaperCT}, with ICLR 24, COLM 24, COLT 24, ACL 24, ACL 23, and ACL 22 as the selected conferences to cover a diverse set of topics. We randomly sampled 2,000 papers for ICLR 24. Table~\ref{tab:conference_stats} presents the statistics of the acquired elements. In total, we collect more than 600 elements for each category from the analysis of 7,483 papers. From the same pipeline, for ACL, the number of method elements that can be extracted from Gemini decreases over the years, \emph{e.g.}, \textit{noise sensitivity}, \textit{multi-criteria optimization}, and \textit{early exit} that appear in ACL 23 no longer appear in ACL 22. With a similar number of papers analyzed, with a similar number of domain elements, ICLR 24 also covers more themes and method elements compared to ACL 2024. Example elements for other conferences are in the appendix (Table~\ref{tab:human_abc_full})

We note that these elements are merged from the raw elements processed from the papers, which can potentially lead to more elements at a finer granular view, \emph{e.g.}, the element \textit{generalization} is merged from \textit{generalizability}, \textit{domain generalizability}, and \textit{temporal generalization}, which can lead to subtle but crucial changes in the paper story and experiment design. We will open-source the finer-granular elements as well as their visited counts.

\subsection{Discussion}

With all the elements in hand, we can then generate the raw ideas by combining them with the templates. We propose two uses for the resources:
(1) randomly sample a template, \emph{e.g.}, \textit{Compare $c_1$ and $c_2$ in $b_1$ under $a_1$}, and randomly fill in the elements; (2) enumerate the top-visited elements and templates to construct a diverse set of raw ideas to fuel the studies on downstream execution or quality evaluation. We further study the characteristics of the generated ideas before and after LLM rewriting in Section~\ref{sec:experiments}. 

In our current design, we treat each equally in sampling after ranking with their visit counts from the papers. We also note that the statistical features, such as the popularity of an element or selectional preference~\citep{zhang-etal-2019-sp} among elements, can potentially suggest a better sampling process for the raw ideas or disk categorization. For example, we can build a Viterbi-like sampling process considering the selectional preference as the transitive scores. On the other hand, a fine-grained element sampling process can lead to a controllable ideation process, \emph{e.g.}, sampling the frequent elements can potentially increase the relevance to specific conferences, while sampling randomly can potentially increase the diversity of the ideas, which leads to a trade-off between the relevance and diversity.

At the current stage, we build the pipeline with pre-defined disk types and leave the extension of the data mining and ideation pipeline for future work.

\section{Experiment and Analysis}
\label{sec:experiments}

In this section, we take a closer look at the characteristics of the raw ideas from the template combination of the elements from the perspective of (1) differences across conferences; (2) comparison with the generated research ideas from previous work.

\subsection{Differences across conferences}

\begin{figure}[!t]
    \centering
    \includegraphics[clip,trim={0cm 0cm 0cm 0cm},width=\linewidth]{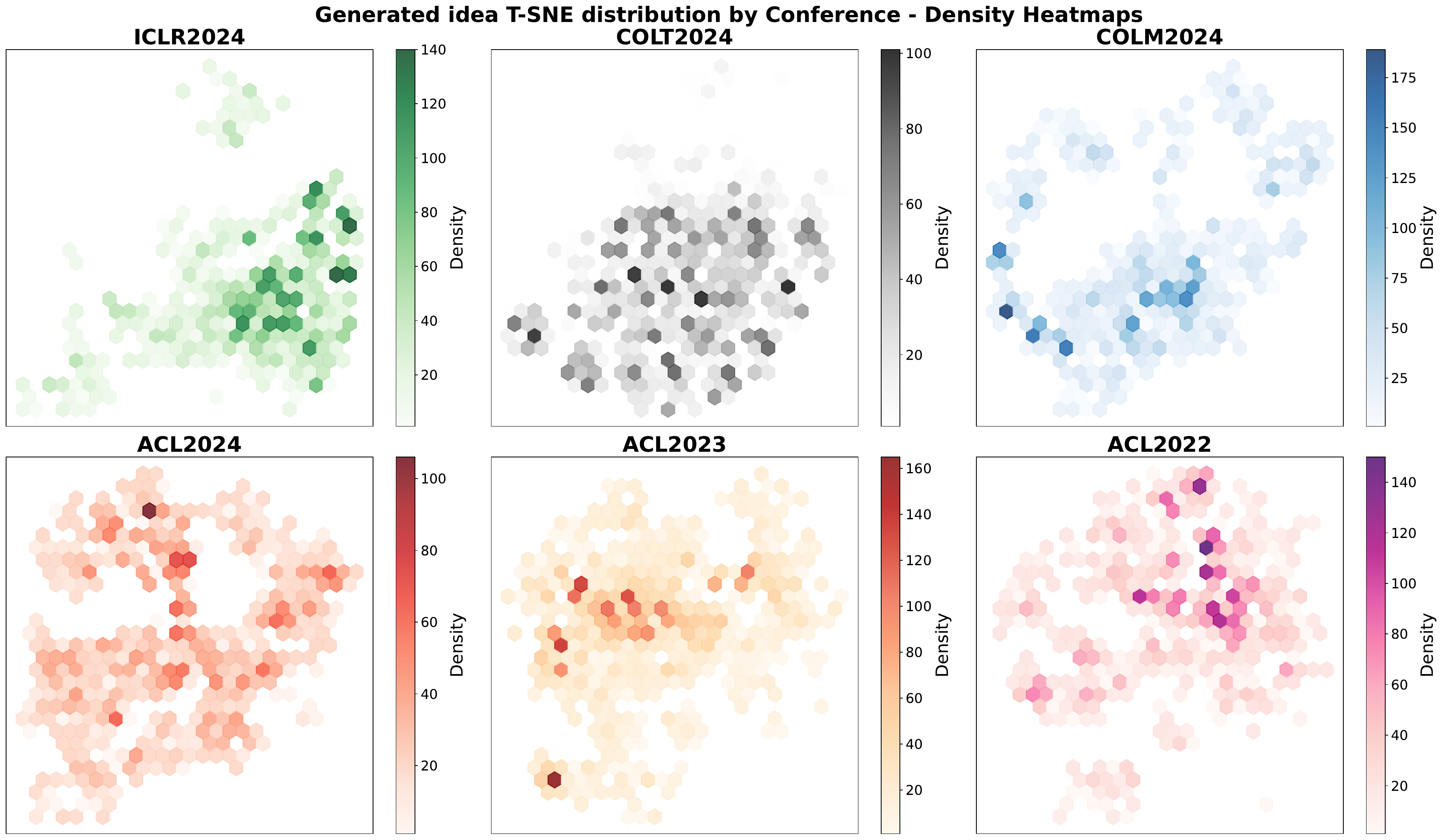} 

\caption{The density heatmap visualization of ideas generated from the basic \textit{A, B, C} template with top-20 most visited elements from each disk for each conference. All sub-figures are aligned in the same distribution with t-SNE. We can observe different relations among the ideas from specific conferences, e.g., taking up different parts of the space.}
    \label{fig:tsne}
\end{figure}

\begin{table}[t]
\centering
\small
\begin{tabular}{p{1.5cm}|p{3.3cm}|p{3.3cm}|p{3.3cm}}
\toprule
\textbf{Conference} & \textbf{Top A (Theme)} & \textbf{Top B (Domain)} & \textbf{Top C (Method)} \\
\midrule
ACL 24 & in-context reasoning, in-the-wild, zero-shot, alignment, benchmarking
& reasoning, question answering, calibration, safety, machine translation, natural language inference
& LLMs, transformers, Self-attention, LoRA, retrieval-augmented generation \\
\midrule
ICLR 24 & generalization, efficiency, robustness, scalability, self-supervised
& reasoning, federated learning, safety, reinforcement learning, planning 
& LLMs, deep learning, transformers, diffusion, vision-language models \\
\bottomrule
\end{tabular}
\caption{Qualitative comparison of the most frequent extracted elements from different conferences. We can observe both shared and different keywords.}
\label{tab:compare_confs_qualitative}
\end{table}

We first compare ideas from different conferences with the extracted element. To avoid the noise from redundant words in the templates, we use the basic \textit{A, B, C} template with the top 20 most visited elements from each disk to generate the raw research ideas. For each conference, we will have 4,000 raw research ideas. Then we convert the research ideas to TF-IDF vectors and apply t-SNE for the visualization.

As presented in Figure~\ref{fig:tsne}, we can observe different relations among the conferences: (1) COLT 2024 ideas (\textcolor{gray}{gray} dots) are comparatively 
standalone, with limited coverage with other conferences. ICLR 2024 ideas (\textcolor{teal}{green} dots) have an overlap with ACL 2024 ideas, but still have a standalone area of clusters; (2) COLM 2024 ideas (\textcolor{blue}{blue} cross) lie in the intersection of ACL 2024, ICLR 2024, and COLT 2024, which shows the joint interests of language modeling from different communities, as the full name of COLM is Conference on
language modeling; (3) As a sanity check, ACL ideas across years are largely overlapping on the upper area, although shifts in interests still can lead to standalone clusters. We present an extended study on the differences of elements of ACL over the years in Appendix~\ref{sec:diff_over_years}. 

We further qualitatively compare the elements for ACL 2024 and ICLR 2024 in Table~\ref{tab:compare_confs_qualitative}, which indicates the causes of the geometrical relations in the t-SNE visualizations: how researchers submit to different conferences show shared (\emph{e.g.}, LLMs and Transformers) and divergent interests (natural language inference vs. federated learning).

\subsection{Comparing different ideation methods}

\begin{table}[t]
\centering
\small
\begin{tabular}{l|c|c|c|c|c}
\toprule
\textbf{Ideation Methods} & \# Ideas & \# Words & Diversity & Similarity & Relevance \\
\midrule
\citet{si2024llmideas} & 93 & 1,063 & 0.29 & 0.22 & 0.28\\
\citet{yu2024researchtown} & 100 & 2,379 & 0.29 & 0.19 & 0.18\\
\midrule
Ram\'{o}n Llull (Top) & 100 & 1,014 & 0.21 & 0.26 & 0.11  \\
Ram\'{o}n Llull (Random) & 100 & 1,105 & 0.41 & 0.23 & 0.05  \\
\bottomrule
\end{tabular}
\caption{Statistics and metric results of different automatic ideation methods. The computation of the similarity and relevance uses ACL 2025 main paper titles as references.}
\label{tab:ideation_methods}
\end{table}

With the elements in hand, we can then generate the research ideas with LLMs rewriting. Specifically, we use Gemini-1.5 Pro to rewrite the sampled combination, for example, with element \textit{emergent}, \textit{theory of mind}, and \textit{variational inference}, the rewritten generated idea is: \textit{Emergent Theory of Mind in Disentangled Latent Spaces via Variational Inference}. We consider two variants of our thinking machine based on the sampling methods: (1) Ram\'{o}n Llull (Top): we select the most visited elements from the disks and enumerate all the combinations; (2) Ram\'{o}n Llull (Random): we randomly sample elements from all disks and ensure that each element only appear once at most.

We compare these rewritten ideas from previous work on ideation: (1) \citet{si2024llmideas} carefully sample and filter AI-generated ideas and list 93 high-quality ideas on 7 NLP topics, including Bias, Coding, Safety, Multilingual, Factuality, Math, and Uncertainty. These ideas are then used for novelty evaluation~\citep{si2024llmideas} and execution study~\citep{si2025ideationexecutiongapexecutionoutcomes}; (2) \citet{yu2024researchtown}
simulate the diverse discussion in the research community and generate ideas in a question-and-answer format. To allow fair comparison. We select 100 ideas from \citet{yu2024researchtown} from the batch where discussions are based on certain previous papers. 
To allow fair comparison, we select 100 ideas from each of our thinking machine variants with elements extracted from ACL 2024\footnote{The authors note that the comparative study is mainly set up to compare the characteristics of these ideation methods. Since the previous methods are designed to sample ideas for their own purposes: novelty evaluation~\citep{si2024llmideas} and research community simulation~\citep{yu2024researchtown}. The results from our metrics do not suggest the superior quality of any method. Rigorous idea quality evaluation may involve extensive expert annotation and insights from execution~\citep{si2025ideationexecutiongapexecutionoutcomes}.}. We only compare the sampled research idea titles for all the methods.

We consider various metrics to compare the ideation methods, including diversity, similarity, and relevance to certain conference papers.

For diversity, we follow~\citep{li2015diversity} to use \textit{distinct-1} as a metric, the number of distinct unigram count normalized by the total number of words to capture the diversity of concise titles. For future work involving abstract or sections, the metric of diversity can be extended to entropy-based metrics as described in ~\citet{zhang2025noveltybench}.

For relevance and similarity, we consider using paper titles from the main track accepted papers from ACL 2025 to ground the comparison, where the accepted papers are released after June 2025. To reduce the chance of LLMs seeing the paper titles in their training data (Gemini 1.5 Pro was released Feb 2024). Specifically, for relevance, we compute the average BLEU score~\citep{papineni2002bleu} between each generated idea and each conference paper title pair to measure how likely a generated title is relevant to the conference. For similarity, we measure how likely a research paper title in ACL 2025 is similar to a generated idea. Similar to our experiments in Appendix~\ref{sec:diff_over_years}, we use token-level Jaccard similarity to capture the similarity of a pair of titles. We report the similarity as the average across ACL 2025 paper titles that scored the top-K highest similarities, where K equals the number of model-generated ideas.

Table~\ref{tab:ideation_methods} presents the different characteristics of different automated ideation methods: Ram\'{o}n Llull (Top) that enumerates combinations of the most trending elements achieved the highest similarity and Ram\'{o}n Llull (Random) that samples random elements achieve the highest diversity, with a decrease in relevance - which demonstrates a trade-off between diversity and similarity/relevance. Our Ram\'{o}n Llull thinking machines also show lower relevance compared to human filter ideas~\citep{si2024llmideas} or ideas from simulated discussions grounded on certain papers~\citep{yu2024researchtown}. One potential reason can be that although the random sampling of elements leads to a diverse set of ideas, they are not necessarily the direction of research acknowledged by the community. Future fine-grained sampling identifying the relations among the elements can be a future direction to improve the ideation process of our Ram\'{o}n Llull thinking machine.


\subsection{Analysis: How much of research ideation is combinatorial?}
\label{sec:coverage}

In this section, we test to what degree the ideation of machine learning research can be explained by our proposed Ram\'{o}n Llull system.
To this end, we conduct a \textit{coverage analysis}. This evaluation tests two complementary aspects:

\begin{itemize}[labelwidth=*,leftmargin=1.3em,align=left, topsep=0pt, nosep]

\item {\bf Decomposition:} Given a research paper title, can it be decomposed into constituent A, B, C elements from our extracted disks? We consider a research idea decomposable if Gemini 2.0 Flash successfully converts the paper title into theme, domain and method elements that our method already extracted.

\item {\bf Reconstruction:} Given the theme, domain and method elements alone, can they be combined to approximately reconstruct the title of the original paper? We consider the research idea reconstructible if Gemini 2.0 can propose a title that is highly similar ($\ge$ 30\% Jaccard similarity) given the extracted keywords.%
\footnote{Additional evaluation details (e.g., prompts) are reported in \ref{appendix:bijective_coverage}.}

\end{itemize}


\begin{table}[t]
\centering
\resizebox{0.95\textwidth}{!}{%
\small
\begin{tabular}{l|c|c|c|c|c|c|c}
\toprule
\textbf{Metric} & \textbf{ACL 2024} & \textbf{ACL 2023} & \textbf{ACL 2022} & \textbf{ICLR 2024} & \textbf{COLM 2024} & \textbf{COLT 2024} & \textbf{Overall} \\
\midrule
Decomp. \% & 99.5\% & 99.2\% & 99.2\% & 99.6\% & 100.0\% & 100.0\% & \textbf{99.5\%} \\
Recon. \% & 17.6\% & 17.8\% & 19.9\% & 15.6\% & 11.0\% & 8.2\% & \textbf{16.4\%} \\
\bottomrule
\end{tabular}%
}
\caption{Bijective coverage results across conferences.}
\label{tab:bijective_coverage}
\end{table}

Table~\ref{tab:bijective_coverage} presents our bijective coverage results across 7,483 papers from six major conferences. We find \textbf{near-universal decomposability (99.5\%):} almost all research papers can be decomposed into our A+B+C framework.
This validates that the three-disk design captures fundamental aspects of research ideation across machine learning communities.

On the other hand, \textbf{reconstructibility is limited (16.4\%)}. That said, a non-trivial fraction of real paper ideas can already be faithfully reconstructed by combining ideas in our proposed Ram\'{o}n Llull framework.
While the framework successfully captures structural building blocks that are nearly universal across machine learning research, we note that the specific instantiation of a research idea may still require creativity and insights that go beyond mere combination of past ideas, and researchers' prior and taste may remain essential to effectively navigate this combinatorial space.

\subsection{Analysis: What is not covered by A+B+C?}
\label{sec:beyond_abc}
Beyond the disk view of automated ideation, we identify other dimensions of the problem as follows, to serve as potential motivations for the community: 

\begin{itemize}[labelwidth=*,leftmargin=1.3em,align=left, topsep=0pt, nosep]
\item \textbf{Perturbation.} Given the same set of A, B, C, the final paper can be drastically different. Akin to the trivial and non-trivial perturbation ($x^2 \rightarrow x^4 \text{vs.} x^2 \rightarrow x^{-1}$) discussed in~\citet{huang2025math}, certain perturbations can potentially change the problem fundamentally. Studying pairs of ideas that have identical elements discovered in our pipeline can potentially allow a fine-grained study on the sparks of non-trivial perturbations that largely reshape the problem.

\item \textbf{The 4th Axis.} In Section~\ref{sec:methodology}, we build our  ideation pipeline with Theme (A), Domain (B), and Method (C) as the disks. Another axis of the machine can be ``algorithms'' vs ``analysis''. Axis C currently says that the research idea should \textit{use} the specified method, but a method can be both \textit{used} and \textit{studied/analyzed}. With this fourth disk, the machine can cover further analysis-based work, such as adversarial examples~\citep{szegedy2013intriguing} and edge-of-stability~\citep{cohen2021gradient}. Such analysis work often leads to the discovery of new/revived phenomena, such as Agreement-on-the-line~\citep{baek2022agreementontheline}, Grokking~\citep{power2022grokking}, and Model Collapse~\citep{shumailov2024ai}. 

\item\textbf{Negation.} Another dimension of non-A+B+C ideas is the negation of commonly believed A+B+C, which often leads to wide community discussion, rethinking of the directions, and improved evaluation, such as the mirage of a phenomenon~\citep{schaeffer2023are}, the gap between automatic and human evaluation~\citep{durmus-etal-2022-spurious,gehrmann2022repairingcrackedfoundationsurvey}, and the misuse of certain tools~\citep{grusky2023rogue}
\end{itemize}

\section{Discussion and Future Work}
In this paper, we propose to create a modern Ram\'{o}n Llull thinking machine for automated ideation, which serves as a lightweight and interpretable tool to create a diverse set of LLM-generated ideas, as well as a perspective to study the commonality and differences in the human ideation process across different communities.
We discuss the intended usage and future work as follows:

\paragraph{Intended Usage.} The proposed Ram\'{o}n Llull thinking machine is \textbf{NOT} intended to (1) conduct DDOS (Distributed Denial of Service) on the current brittle reviewing system~\citep{Kim2025PositionTA}; (2) evaluate or attack certain human-generated ideas. Our Ram\'{o}n Llull's Thinking Machine is intended to serve as (1) a baseline for future study on ideation with a filtered set of components, i.e., theme, topics, and domains; (2) motivation for human researchers to track the field status quo and their own ideas. We plan to open-source 1,000 high-quality human-filtered ideas to conduct a stealth human study on the execution of the ideas in a human-AI collaborative manner with real research labs.

\paragraph{Future Work.} We expect to extend our pipeline through (1) evaluating the quality of the generated ideas; (2) studying the human ideation and polishing process through the lens of Ram\'{o}n Llull's Thinking Machine; (3) studying how the execution process can serve as elements or factors of sampling.



\clearpage
\section*{Acknowledgments}
The authors thank Hongming Zhang, Sihao Chen, Zhiyuan Zeng, Wenhao Yang, Fengyu Cai, and Ben Zhou, 
as well as other fellow AI2 interns (including but not limited to Hita, Yapei, Fede, Anej, Amanda, Michael, Nishant, Peiling, Alexiss, and etc) and UW students, for their insights into design and evaluation choices. The authors also thank the constructive discussions with colleagues from CMU WInE Lab.
Xinran Zhao is supported by the ONR Award N000142312840.
This work is supported by the OpenAI Research Credit program, the Amazon AI Research Gift Fund, and the Gemma Academic Program GCP Credit Award. Any opinions, findings, and conclusions or recommendations expressed in this material are those of the authors and do not necessarily reflect those of the sponsors.

\section*{Ethics Statement}
We foresee no ethical concerns or potential risks in our work.
All of the datasets are open-sourced and from peer-reviewed research papers, as shown in Section~\ref{sec:elements}. 
The LLMs we applied in the experiments are also publicly available.
Given our context, the outputs of LLMs are unlikely to contain harmful and dangerous information. The experiments in our paper are mainly on English.

\bibliography{colm2025_conference}
\bibliographystyle{colm2025_conference}

\clearpage

\appendix
\section{Appendix}

\subsection{Limitations}
\label{appendix:limitations}

\paragraph{Evaluating idea quality and novelty.} 
Our current evaluation is based on quantitative metrics such as diversity and community relevance.
While these metrics are useful for assessing the breadth and community-alignment of the generated idea space, they could be insufficient for judging the scientific merit of individual idea under some circumstances.
For instance, a generated idea that is lexically unique in terms of diversity, but could be conceptually trivial or scientifically unsound.
An idea might achieve high relevance by closely mirroring existing research trends, making it plausible but potentially incremental and not truly novel. Conversely, a truly groundbreaking idea might score low on relevance because it deviates significantly from established paradigms.
A more rigorous assessment requires moving beyond surface-level statistics to semantic evaluation, for which human expert judgment remains the gold standard.
Experts assess ideas along critical aspects like feasibility, potential impact, and non-obviousness, providing qualitative depth that text-based metrics are not designed to capture.
However, large-scale human evaluation is difficult to scale, expensive, and subject to inter-annotator variability. Securing a diverse pool of experts capable of judging ideas across the wide range of generated topics is a major logistical challenge.
LLM-as-a-judge frameworks, such as the one proposed by \citep{alpaca_eval} trained with carefully designed rubrics, might be biased by its training data,
potentially favoring well-phrased but shallow ideas over more bad-worded but conceptually deep ones.
A future direction is a hybrid evaluation pipeline that leverages our quantitative metrics for initial filtering, employs LLM-as-a-judge for scalable scoring, and incorporates targeted human experts for final validation.


\paragraph{Organizing the elements.} 
While our current implementation, which treats conceptual elements as independent items, has successfully generated a wide breadth of ideas, it can be further enriched by the structured relationships that exist between scientific concepts.
One extension is to evolve from simple, flat lists to categorical and hierarchical structures by linking to the keywords in OpenReview or the task and method hierarchies on Papers with Code\footnote{\url{paperswithcode.com}}.
This would enable more granular control over ideation — for example, allowing sampling at different level of abstraction.
Another possible direction is to explicitly model the exclusiveness of selection preferences to learn which combinations of themes, domains, and methods are most likely to be coherent.
By integrating such structured knowledge, our system would transform into a more semantic-aware  ideation that is capable of generating ideas that are both novel and conceptually sound.


\subsection{Extended Discussion}
\label{appendix:extended_discussion}

\paragraph{Related Work: Data Mining from Literature} Extracting structured information from academic papers is a critical research area~\citep{zhang2024automated}. Prior work has explored concept-level understanding through methods such as topic discovery~\citep{lee2022taxocom} and concept matching~\citep{kang2024taxonomy}, often operating over large sets of concepts using clustering or taxonomy construction. Ontology-based approaches~\citep{muller2004textpresso} similarly aim to organize and retrieve scientific knowledge from literature at the conceptual level. In contrast, our work focuses on identifying a small set of high-quality concepts, specifically, the theme, domain, and method of each article, that are used as rotating wheels in Llull's thinking machines. This design supports combinatorial exploration and enables ideation within and across research communities.

\paragraph{How to view A+B+C?} The same A+B+C can lead to different results and execution. For the utility and feasibility of the idea execution, it is vital to analyze why the components are complementary: \emph{e.g.}, why a core problem in a domain requires an architectural change or can be viewed as a specific theme.

\subsection{Mining the elements and templates (Full)}
\label{appendix:elements}

\begin{table}[t]
\centering
\small
\begin{tabular}{p{1.8cm}|p{4cm}|p{3cm}|p{3cm}}
\toprule
\textbf{Community} & \textbf{A (Theme)} & \textbf{B (Domain)} & \textbf{C (Method)} \\
\midrule

ICLR 24 & representation learning, offline learning, sparsity, interpretability, explainable, unsupervised learning, uncertainty, multi-modal, multi-hop, reference free, fairness, contrastive learning, sampling, heterogeneity, out-of-distribution, active learning,  hierarchical structure, meta-learning & planning, safety, reinforcement learning, question answering, calibration, automated research, federated learning, classification, image generation, optimization, memorization, representation learning, segmentation & Ordinary Differential Equations, visualization tool, dynamic programming, matching function, plug-in modules, semi-supervised learning, self-training, Text-to-Image Generators, Linear Discriminant Analysis\\
\midrule
COLM 24 & in-context learning, in-the-wild, compositionality, self-evolve, long-tail, multi-hop, reference free, multi-modal, generalization, alignment, adaptation, robustness, granularity, multilingual, interpretability & inference, RAG, automated translation, decision-making, drug discovery, text generation, fine-tuning, argument mining, code editing, preference learning & Transformers, Self-attention, Mamba, RWKV, SSMs, state space models, RLHF, PPO, Mixture-of-Experts, LoRA, RNNs, VQAs, Pruning, deep generative models \\
\midrule
ACL 24 & self-evolve, generalization, domain generalization, temporal generalization, robustness, resilient, parameter-efficient, multilingual, cross-lingual, multi-task learning, cross-task, bias, debiasing, modularity, less is more, adaptive, adaptability, unsupervised adaptation & human-bot interaction, entity grounding, finance, equity research, macroeconomics, tool learning, metaphor interpretation, game playing, open-world games, style transfer, medical diagnostics &  RoBERTa, BART, ByT5, diffusion models, Latent Diffusion Model, Brownian Bridge process, PLMs, Spiking Neural Network, generative models, contrastive decoding  \\
\bottomrule
\end{tabular}
\caption{\small{Lists of Theme (A), Domain (B), and Method (C) written by researchers from different communities. NLP, CV, and RL Theory denote natural language processing, computer vision, and reinforcement learning theory, respectively.}}
\label{tab:human_abc_full}

\end{table}

We present the full table of elements written by humans in Table~\ref{tab:human_abc_full}.

\subsection{Differences over years}
\label{sec:diff_over_years}

\begin{table}[t]
\centering
\small
\begin{tabular}{l|c|c|c}
\toprule
\textbf{Jaccard.} & ACL 22 vs. 23 & ACL 23 vs. 24 & ACL 22 vs. 24\\
\midrule
\# Theme (A) & 0.08 & 0.07 & 0.14 \\
\midrule
\# Domain (B) & 0.22 & 0.17 & 0.19 \\
\midrule
\# Method (C) & 0.05 & 0.08 & 0.09  \\
\bottomrule
\end{tabular}
\caption{\small{Jaccard similarity of different disks for different years of ACL conferences. Compared to theme and method, there is a higher similarity of domains across years.}}
\label{tab:acl_shift_over_years}
\vspace{-0.1in}
\end{table}

Besides the relevance among conferences, another interesting dimension is the distribution shift over years~\citep{tran2020openreviewopenreviewcritical}. We further compare the elements of different disks through the lens of token-level Jaccard similarity for ACL from 2022 to 2024. Table~\ref{tab:acl_shift_over_years} shows a higher similarity in the elements from domains compared to themes or methods. One potential reason is that ACL typically lists several tracks to guide the paper submission, \emph{e.g.}, \textit{Question Answering} and \textit{NLP applications}. The Jaccard similarity does not change a lot across years, but the similarity is not high in general, which indicates the topical diversity in top-tier conferences. Besides the disappearance of method elements through the years, we can also observe the occurring interests in certain elements. For the elements that appear uniquely in ACL 2024,  disk A (theme) has \textit{perspective awareness}, \textit{Multi-generator}, etc; disk B has \textit{Hateful Meme Detection}, \textit{emotional support}, etc.

\subsection{Example ideas from different ideation methods}
This section provides examples from three distinct AI-driven ideation methodologies, each producing a different kind of conceptual output. We summarize the idea and omit some details for better presentation.  
\clearpage

\subsubsection{Example 1: AI Scientist}
This method demonstrates the LLM's capacity to generate a comprehensive, structured research plan from a single core concept. The output is an actionable roadmap detailing the necessary steps to investigate an idea for next-stage experimentation.

\begin{tcolorbox}[fit,height=20cm,interior hidden, width=\linewidth,borderline={1pt}{-2pt}]
\textbf{Title:} Adversarial Stereotype Dissolution Prompting: Reducing Social Biases in Large Language Models through Active Counter-Example Generation\\

\textbf{1. Problem Statement:} 
Large language models often generate outputs that reinforce existing stereotypes and social biases, even when attempting to be unbiased. This perpetuates harmful societal prejudices and limits the models' ability to provide fair and inclusive responses across diverse user groups.\\

\textbf{2. Motivation:} Current approaches to reducing bias in language models typically focus on avoiding or counterbalancing stereotypes... By prompting the model to generate adversarial examples that contradict stereotypes, we can encourage it to develop more nuanced and less biased representations ...\\

\textbf{3. Proposed Method:} 
We introduce \textbf{Adversarial Stereotype Dissolution Prompting (ASDP)}, a technique that challenges the model to actively generate counter-stereotypical examples. The prompt structure includes: ...\\

\textbf{4. Step-by-Step Experiment Plan:}
\begin{enumerate}[label=\textbf{Step \arabic*:}, leftmargin=*]
    \item \textbf{Dataset Preparation}:\\ Create a dataset of stereotype-sensitive queries across various domains (e.g., gender, race, age, profession), Collect 100-200 such queries for a comprehensive evaluation...

    \item \textbf{Baseline Methods Implementation}:\\ Implement the following baseline methods:
        \begin{enumerate}[label=\alph*)]
            \item Standard prompting (direct query).
            \item Disclaimer prompting (adding ``Please provide an unbiased response'' to queries).
            \item Counterbalancing prompting (explicitly asking for examples from different groups).
        \end{enumerate}

    \item \textbf{ASDP Implementation}
    \begin{itemize}
        \item Implement the Adversarial Stereotype Dissolution Prompting method.
        \item Create a template that includes the four steps mentioned in the proposed method.
        \item Ensure the prompt is clear and consistent across different queries.
    \end{itemize}

    \item \textbf{Model Selection}
        Use GPT-4 and GPT-3.5-turbo from OpenAI's API for the experiments. These models are state-of-the-art and widely used, making the results relevant and comparable.

    \item \textbf{Experiment Execution}: For each query in the dataset:\\
        a). Generate responses using each baseline method and ASDP.\\
        b). For ASDP, store the intermediate outputs (identified stereotype, counter-examples, analysis, and reformulated query) for later analysis.
        
    \item \textbf{Evaluation Metrics}: Develop a set of evaluation metrics:\\
        a). \textit{Stereotype Adherence Score:} Manually rate responses on a scale of 1-5 for how much they adhere to common stereotypes.\\
        b).\textit{Diversity Score:} Measure the lexical and semantic diversity of the generated responses.\\
        c).\textit{Factual Accuracy:} Verify the factual claims made in the responses.\\
        d). \textit{Plausibility of Counter-Examples:} Rate the realism and plausibility of the generated counter-examples in ASDP.

    \item \textbf{Human Evaluation}: Recruit a diverse group of 5-10 human evaluators to rate a subset of the responses (50-100) on bias, fairness, and overall quality...

    \item \textbf{Analysis}
    \begin{itemize}
        \item Compare the performance of ASDP against the baselines using the defined metrics.
        \item Analyze the intermediate outputs of ASDP to understand how the model identifies and challenges stereotypes...
    \end{itemize}
\end{enumerate}
...
\end{tcolorbox}

\subsubsection{Example 2: Research Town (Author Simulation)}
This method simulates a domain expert to synthesize a focused and plausible research abstract. By adopting the persona of specific researchers, the system generates a condensed, high-impact summary of a potential scientific contribution.
\begin{tcolorbox}[fit,height=5cm,interior hidden, width=\linewidth,borderline={1pt}{-2pt}]
\textbf{Seed Idea:} Language Models as Memory Augmentation \\
\textbf{Simulated Authors:} Alexei A. Efros, David A. Forsyth

\textbf{Generated Idea:}
\begin{itemize}
  \item title: Language Models as a Cognitive Prosthesis for 
           Memory Augmentation",
  \item authors: "Alexei A. Efros, David A. Forsyth,
  \item abstract: Human memory is fallible... We propose a system, 
             'Cognitive Scribe,' that leverages large 
             language models (LMs) to serve as a personal 
             memory prosthesis...,
  \item method: Our proposed method involves three main components: 
             (1) A wearable device... (2) A secure, on-device 
             data processing pipeline... (3) A fine-tuned large 
             language model...,
  ...
\end{itemize}
\end{tcolorbox}

\subsubsection{Example 3: Thinking Machine}
This method is designed to produce foundational concepts that can define new avenues of inquiry. The output is typically a concise, high-level idea that represents a strategic direction rather than a detailed plan.
\begin{tcolorbox}[width=\linewidth, colback=white, colframe=black, boxrule=1pt]
\textbf{Generated Idea Title:}
\begin{quote}
\textit{Evolving Research Agents: Autonomous Iteration of Hypotheses, Experiments, and Refinement}
\end{quote}
\end{tcolorbox}
Instead of detailing a solution to a known problem, it proposes the creation of a \textit{Evolving Research Agents} that autonomously conduct science.
The output idea is a straightforward, strategic concept, suggesting a possible and easy paradigm for how LLM can assist the process of idea discovery.

\clearpage
\subsection{Experimental Details}
\label{appendix:experimental_details}

For Gemini, Claude, and GPT models, we use the official API service.  If applicable, we set the max output token to be 8192, temperature to be 0.7, top p to be 0.7, and top k to be 50. For TF-IDF and t-SNE, we use the implementations of Scikit-Learn.

We present the details of prompts for element extraction and element merging as follows:

\begin{tcolorbox}[fit,height=14cm,interior hidden, width=\linewidth,borderline={1pt}{-2pt}]

\textbf{Element Extraction Prompt}

You are a helpful assistant who annotates the paper with its title and the abstract:

Please annotate the paper with the following information:

1. The themes of the paper (As, \emph{e.g.}, few-shot, long-tail, less is more, in-the-wild, self-refine, look-ahead, hindsight, memory, self-, rethink, weak to strong, granularity, in-context learning, reference free, grokking, self-evolve, long-tail, compositionality, multi-hop, modular, etc.)
2. The domains of the paper (Bs, \emph{e.g.}, question answering, argument mining, planning, RAG, calibration, reasoning, safety, debate, memorization, automated research, etc.)
3. The method insights of the paper, especially novel architecture (Cs, \emph{e.g.}, Mamba, RWKV, LLMs, Self-attention, LLMs, etc.)
4. The templates of the paper title/abstract (templates, \emph{e.g.}, Comparing C1 and C2 in B1 with A1, etc.)

Requirements:
1. There can be multiple A, B, C, and one Template. 
2. Use generic keywords of A, B, C, and Template to allow reuse, instead of specific ones for each paper.
3. Make sure keywords are exclusive among A, B, C.

Please output the annotation in the following JSON format:

{"A": ["few-shot", "long-tail"], "B": ["argument mining"], "C": ["Mamba"], "Template": ["Comparing C1 and C2 in B1 with A1"]}

An Example:
Title: Thrust: Adaptively Propels Large Language Models with External Knowledge

Abstract: Although large-scale pre-trained language models (PTLMs) are shown to encode rich knowledge in their model parameters, the inherent knowledge in PTLMs can be opaque or static, making external knowledge necessary. However, the existing information retrieval techniques could be costly and may even introduce noisy and sometimes misleading knowledge. To address these challenges, we propose the instance-level adaptive propulsion of external knowledge (IAPEK), where we only conduct the retrieval when necessary. To achieve this goal, we propose to model whether a PTLM contains enough knowledge to solve an instance with a novel metric, Thrust, which leverages the representation distribution of a small amount of seen instances. Extensive experiments demonstrate that Thrust is a good measurement of models' instance-level knowledgeability. Moreover, we can achieve higher cost-efficiency with the Thrust score as the retrieval indicator than the naive usage of external knowledge on 88\% of the evaluated tasks, with 26\% average performance improvement. Such findings shed light on the real-world practice of knowledge-enhanced LMs with a limited budget for knowledge seeking due to computation latency or costs.

Output: \{"A": ["adaptive"], "B": ["RAG"], "C": ["Large Language Models"], "Template": ["A1 application of B1 to C1"]\}

You task:

Title: {title}

Abstract: {abstract}

Output:               
\end{tcolorbox}

\begin{tcolorbox}[fit,height=5cm,interior hidden, width=\linewidth,borderline={1pt}{-2pt}]
\textbf{Element Merging Prompt}

You are a helpful assistant who merges the keywords or phrases with their semantic similarity.

Here is a list of keywords or phrases for a {domain}:
{keywords}

Requirements:
1. Please merge the keywords by creating a keyword group in a valid decodable JSON format.
2. No need to merge the keywords that are not to similar.
3. Output the JSON format only.
4. Do not be lazy, please list the full output covering all keywords or phrases without omission.

The potential JSON format is:
\{\{"keyword\_group\_name": ["keyword1", "keyword2", "keyword3"]\}\}

The keyword group name should be a short and concise description of the keyword group. An example keyword group: "RAG": [RAG,retrieval augmented generation, retrieval augmentation]

Your output:             
\end{tcolorbox}

\begin{tcolorbox}[fit,height=3cm,interior hidden, width=\linewidth,borderline={1pt}{-2pt}]
\textbf{Idea Rewriting Prompt}

You are a senior professor in AI, and your students propose to do a combination. Can you refine the title into a good one that can be accepted by top conferences such as ACL 2025 and ICLR 2026? Please output one title only, with no other text. Requirements: 1. Do not hallucinate, 2. do not use any existing paper names in your pretraining data. 3. make sure the title is with an outstanding paper quality so that your student can be happy and successfully graduate.           
\end{tcolorbox}

\subsection{(Details) Elements mined from conferences}
\label{appendix:conf_abc}

\begin{table}[t]
\centering
\small
\begin{tabular}{p{1.8cm}|p{4cm}|p{3cm}|p{3cm}}
\toprule
\textbf{Community} & \textbf{A} & \textbf{B} & \textbf{C} \\
\midrule
ACL 2024 & adaptive, less is more, hierarchical, in-the-wild, self-refine, hindsight, rethink, grokking, long-tail, compositional, multi-hop  & agent, planning, retrieval, safety, calibration, reasoning, memorization, persuasion, debate & Mamba, RL, Linear Models, KV Cache, Quantization, Diffusion, Self-attention, Self-supervision \\
\midrule
CV & test-time Training, meta learning, active learning, open-set calibration, open-vocab grounding, continual learning, knowledge guided learning, inverse rendering & image classification, detection, segmentation, optical flow estimation, action recognition, style-transfer, denoising & vision transformer, NeRF, ConvNext, style-GAN, point-transformer, Perceiver, Instant-NGP, Yolo, UNet, LoRA\\
\midrule
RL Theory & Value A3 & Value B3 & Value C3 \\
\bottomrule
\end{tabular}
\caption{Lists of Theme (A), Domain (B), and Method (C) written by researchers from different communities. NLP, CV, and RL Theory denote natural language processing, computer vision, and reinforcement learning theory, respectively.}
\label{tab:conf_abc}
\end{table}

\subsection{Bijective Coverage Evaluation Details}
\label{appendix:bijective_coverage}

For the bijective coverage analysis in Section~\ref{sec:coverage}, we implement a two-stage evaluation process using Gemini 2.0 Flash.

\paragraph{Decomposition Prompt.} We use the following prompt to test whether research papers can be decomposed into our A+B+C framework:

\begin{tcolorbox}[fit,height=5cm,interior hidden, width=\linewidth,borderline={1pt}{-2pt}]

You are an expert in AI research taxonomy. I will give you lists of research themes (A), domains (B), and methodologies (C), and a paper title. Your task is to find the MOST SPECIFIC and ESSENTIAL concepts from these lists that capture the core of this paper.

THEMES (A): \{themes\}

DOMAINS (B): \{domains\}

METHODOLOGIES (C): \{methodologies\}

PAPER TITLE: "\{title\}"

Extract the most essential concepts that would allow someone to reconstruct a similar title:
- Select relevant themes from list A
- Select relevant domains from list B  
- Select relevant methodologies from list C

Focus on concepts that are ESSENTIAL to the paper's contribution, not just tangentially related.

Respond with a JSON object:
\{\{"selected\_A": ["theme1", "theme2"], "selected\_B": ["domain1"], "selected\_C": [``methodology''], "confidence": 0.0-1.0, "explanation": "brief explanation"\}\}
\end{tcolorbox}

\paragraph{Reconstruction Prompt.} For testing reconstruction capability, we use:

\begin{tcolorbox}[fit,height=3cm,interior hidden, width=\linewidth,borderline={1pt}{-2pt}]

You are a senior AI researcher. Given these research concepts, generate 5 different realistic paper titles that combine them:

THEMES: \{themes\}
DOMAINS: \{domains\} 
METHODOLOGIES: \{methodologies\}

Generate 5 diverse paper titles that would be suitable for a top-tier conference like ACL/EMNLP/NeurIPS. Each title should:
1. Combine all the given concepts naturally
2. Sound like a real research paper title
3. Be specific and technical
4. Be different from the others

Format as a numbered list:
1. [Title 1]
2. [Title 2]  
3. [Title 3]
4. [Title 4]
5. [Title 5]
\end{tcolorbox}

\paragraph{Evaluation Metrics.} We consider a paper \textit{decomposable} if it can be successfully mapped to at least one element from each disk (A, B, C). For \textit{reconstruction}, we generate 5 candidate titles and compute Jaccard similarity between each candidate and the original title, taking the maximum similarity. Papers with similarity $\geq$ 30\% are considered successfully reconstructible.

\subsection{Original Ram\'{o}n Llull’s \textit{Ars combinatoria}}
\label{appendix:original_machine}

\begin{figure}[!t]
    \centering
    \includegraphics[clip,trim={0cm 0cm 0cm 0cm},width=0.7\linewidth]{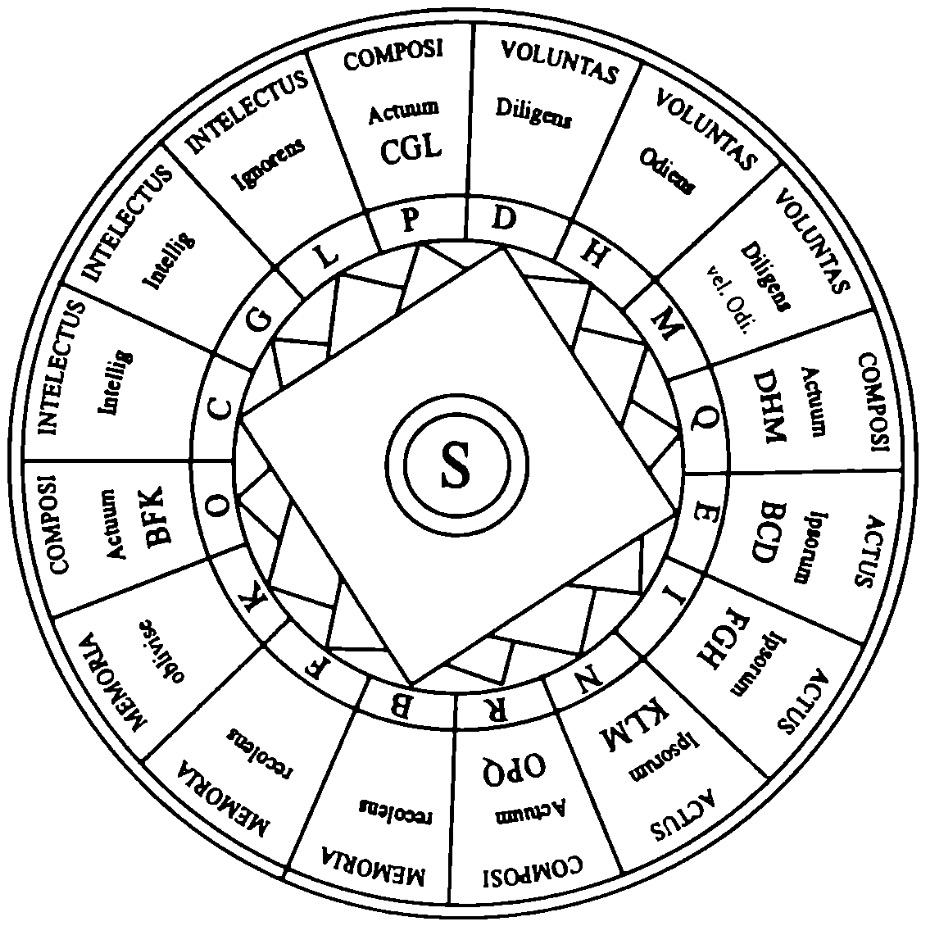} 

\caption{\small{The illustration of the original Ram\'{o}n Llull’s thinking machine.}}
    \label{fig:original_machine}
  \end{figure}

We present the original Ram\'{o}n Llull’s thinking machine in Figure~\ref{fig:original_machine} from \citet{borges1937}.

\end{document}